%% file: acl.tex
\pdfoutput=1

\documentclass[11pt]{article}

\usepackage[final]{acl}

\usepackage{times}
\usepackage{latexsym}

\usepackage[T1]{fontenc}

\usepackage[utf8]{inputenc}

\usepackage{microtype}

\usepackage{inconsolata}

\usepackage{graphicx}

\usepackage{enumitem}
\usepackage{flushend}
\usepackage[linesnumbered,ruled,vlined]{algorithm2e}

\usepackage{amsmath}
\usepackage{subcaption}
\usepackage{stfloats}
\usepackage{float}
\usepackage{booktabs}

\usepackage[useregional]{datetime2}

\usepackage{arydshln}

\newcommand{\gptdate}{\DTMdisplaydate{2024}{1}{24}{-1}}

\title{ReALM: Reference Resolution As Language Modeling}

\author{Joel Ruben Antony Moniz\thanks{\enspace Equal contribution} $^1$, Soundarya Krishnan\footnote[1]{}$\;^2$, Melis Ozyildirim${}^3$,  \\
        \textbf{Prathamesh Saraf, Halim Cagri Ates, Yuan Zhang, Hong Yu${}^4$ }\\
        \{${}^1$joelmoniz, ${}^2$skrishnan22, ${}^3$melisozyildirim, ${}^4$hong\_yu\}@apple.com \\
        Apple}

\begin{document}
\maketitle

\begin{abstract}
Reference resolution is an important problem, one that is essential to understand and successfully handle contexts of different kinds. This context includes both previous turns and context that pertains to non-conversational entities, such as entities on the user's screen or those running in the background. While LLMs have been shown to be extremely powerful for a variety of tasks, their use in reference resolution, particularly for non-conversational entities, remains underutilized. This paper demonstrates how LLMs can be used to create an effective system to resolve references of various types, by showing how reference resolution can be converted into a language modeling problem, despite involving forms of entities like those on screen that are not traditionally conducive to being reduced to a text-only modality. We demonstrate large improvements over an existing system with similar functionality across different types of references, with our smallest model obtaining absolute gains of over 5\% for on-screen references. We also benchmark against GPT-3.5 and GPT-4, with our smallest model achieving performance comparable to that of GPT-4, and our larger models substantially outperforming it.
\end{abstract}

\section{Introduction} 

Human speech typically contains ambiguous references such as "they" or "that", whose meaning is obvious (to other humans) given the context. Being able to understand context, including references like these, is essential for a conversational assistant that aims to allow a user to naturally communicate their requirements to an agent, or to have a conversation with it \cite{luger2016like, ljungholmvoice}. In addition, enabling the user to issue queries about what they see on their screen is a crucial step in ensuring a true hands-free experience in voice assistants.  For instance, consider the following interactions between a user and an agent shown in Table~\ref{tab:validinvalidaugmentation}.

\begin{table}[!htbp]
\small
\centering
\caption{Sample Interactions between a user and an agent.}
\label{tab:validinvalidaugmentation}
\setlength{\tabcolsep}{12pt}
\begin{tabular}{ll}
\toprule
\textbf{Speaker} & \textbf{Dialogue} \\
\midrule
User & Show me pharmacies near me \\
Agent & Here is a list I found. \\
Agent & ... (list presented) \\
User (eg 1) & Call the one on Rainbow Rd. \\
User (eg 2) & Call the bottom one. \\
User (eg 3) & Call this number (present onscreen) \\
\bottomrule
\end{tabular}
\end{table}

Here, it is immediately apparent that it would not be possible for the Agent to understand or complete the user's query without the ability to use and comprehend context. It also stands to reason that there are multiple types of context that are necessary to handle user queries: conversational context, on-screen context, and background entities.

Recent Large Language Models (LLMs) \cite{stammbach-etal-2022-heroes, touvron2023llama, santhanam-etal-2022-colbertv2,dettmers2023qlora} have often enabled end-to-end experiences, perhaps even obviating the need of a traditional multi-stage pipeline that includes reference resolution \cite{khatri2018advancing}. There are, however, still several real-world cases where a pipeline is valuable, perhaps even essential, and an end-to-end approach falls short. 
First, when a framework runs completely on-device (for example, for privacy and efficiency reasons) on a system such as a smartphone that has relatively limited computing power, due to the low-power nature of the system and latency constraints, using a single, large, end-to-end model is infeasible: using a single LLM for this task would usually require the use of a large model with long prompts for true end-to-end experiences \cite{wei2022emergent}.
Second, consider the case when the model has to integrate with APIs, has to consume information from components upstream, or has to provide information to be consumed downstream: while in these cases it is possible to have an end-to-end approach having the LLM write API calls \cite{patil2023gorilla,qin2023toolllm}, this often requires a large language model and a complete overhaul of existing pipelines, which might be cumbersome or completely infeasible.
Third, the use of a focused model would allow for an existing reference resolution module to be swapped with improved versions in a transparent way, while providing improved ability to hill-climb and improved interpretability, by virtue of the system being modular. Finally, for the task under consideration in this paper, reference resolution does not include solely conversational references, but also includes the ability to reference an on-screen and/or a background entity that is part of what the user currently perceives in their interaction with a device, but has not been a part of the conversational history that results from their direct interaction with the virtual agent in question.
There thus continues to be utility in exploring "traditional" NLP tasks such as reference resolution, despite some of the larger language models being able to handle them implicitly. In this work, we thus advocate the use of (relatively) smaller language models, but fine-tuned for specifically and explicitly for the task of reference resolution.

 Along similar lines, relying on language modeling alone \cite{bajaj2022metro,patra2022beyond,zheng2023judging} has recently shown great promise in being able to handle a variety of tasks \cite{wang2018glue,wang2019superglue,hendrycks2020measuring,wei2021finetuned,chung2022scaling}, such as causal reasoning, linguistic acceptability, question answering, textual entailment and even coreference resolution: Using Language Models (LMs) does exceedingly well on tasks that can be modeled in a sequence-to-sequence fashion. However, the biggest challenge with adopting this technique for the general reference resolution task in the context of a voice assistant lies in resolving references to entities on the screen and using their properties, in other words, getting the LM to, informally speaking, “see”. In particular, it is non-obvious how to encode entities on a screen in a manner that is conducive to being resolved by an LM, while also being consistent enough with how conversational entities are encoded to enable the LM to successfully perform reference resolution on both types of entities.

In this work, we propose reconstructing the screen using parsed entities and their locations to generate a purely textual representation of the screen that is visually representative of the screen content. The parts of the screen that are entities are then tagged, so that the LM has context around where entities appear, and what the text surrounding them is (Eg: call the business number).
To the best of our knowledge, this is the first work using a Large Language Model that aims to encode context from a screen.

\section{Realated Work and Motivation}

While traditional reference resolution systems have explored conversational and visual/deictic references in great depth \cite{kottur2018visual,schwartz2019factor,kang2019dual}, resolving on-screen references is a domain that has been relatively under-explored. However, as shown above, conversational agents on a mobile device need to understand references to the screen, and to support such experiences, to be truly natural. 
On screen references differ from visual and deictic references for several reasons: they tend to be more structured and highly textual, which enables the use of a lighter model to treat this as a text-only problem without a visual component; further, user queries around on-screen elements often tend to be more action-oriented rather than QA based; finally, they use synthetic screens rather than natural real-world images, which are much easier to parse, but whose distribution completely differs from that on which larger pre-trained image-based systems (such as CLIP \cite{radford2021learning}) tend to be trained.
Further, jointly being able to perform conversational and on-screen reference resolution has been even less explored, with prior work often focusing on images and graphics \cite{willemsen2023resolving}, or UI elements \cite{you2024ferret}. 

Vision transformers \cite{dosovitskiy2020image, touvron2021training, liu2021swin,yu2021vector} and other pre-trained models have recently gained prominence as a popular first step in tasks that require visual understanding. However, these tend to be trained on natural, real-world images rather than screenshots of on-screen layouts, which have a very different distribution. In addition, these can be extremely expensive to (pre-)train, requiring a very large number of images and several hundred GPU hours (or more). Further, they tend to not perform as well on images heavily embedded with text, and dedicated textual understanding approaches \cite{xu2020layoutlm,xu-etal-2021-layoutlmv2,hwang-etal-2021-cost,hwang-etal-2021-spatial,hong2022bros} tend to heavily rely on multiple modules such as bounding box detection and OCR while also relying on good image quality. Joint vision+text models are also substantially more expensive with respect to parameters and computational cost. Finally, these models would need to parse text to be able to perform function (Eg: ``call the business number'' needs to extract the number associated with the business landline from the raw image), a process which can be complex and compute intensive when bearing in mind that the underlying text and its location on the screen has been referred by the system, and as a consequence can be relatively easily extracted without large, complex models.

The most closely related work which we are aware of, and which we consequently use as our baseline, is that of \citet{ates2023marrs}, an extension of \citet{bhargava2023referring} which deals purely with on-screen references; however, it suffers from several drawbacks, which we address in this work. First, these approaches rely on a dedicated ``Category module'' to deal with type-based references. This module often requires manually on-boarding entities every time a new type is created (a common occurrence in voice assistants, as the supported functionality of the assistant is expanded over time). In addition, such modules often treat each type as distinct, with the similarity of different types ignored. This, in turn, leaves on the table the potential positive transfer that could have happened between semantically related classes (such as ``phone number'' and ``contact'') when data is added for one of those classes. This approach is thus difficult to scale to new entity types and use cases. Second, these systems rely on the use of hand-crafted rule-based textual overlap features, which require heavy feature engineering and tend not to be robust. In addition, these heuristics often do not account for semantic similarity, and are not able to encode real-world understanding or commonsense reasoning. Finally, these methods effectively classify how related each entity is to the query in question independently of all other entities and later threshold them, whereas our current approach directly picks out the most relevant option (or options), while also allowing for no entities to be relevant. Our approach thus additionally has the advantage of removing the reliance on a set threshold, while also providing all the functionality supported in the previous approaches.

\section{Task}
 We formulate our task as follows: 
 Given relevant entities and a task the user wants to perform, we wish to extract the entity (or entities) that are pertinent to the current user query. The relevant entities are of 3 different types:

 \begin{enumerate}[leftmargin=*]
     \item On-screen Entities: These are entities that are currently displayed on a user's screen
     \item Conversational Entities: These are entities relevant to the conversation, which predominantly include those that come from a previous turn. For example, let's say that the first turn of the user is ``Call Mom.'', which is an unambiguous turn that uses a contact called Mom. Shortly after, if the user says ``Text her'', the reference ``her'' needs to be resolved to the contact for ``Mom'' that was brought up in the previous turn; this contact is thus a conversational entity. Another example might involve an interaction in which the user requests for a list of places or alarms to choose from (or the agent presents one for a user turn such as ``Show me pharmacies near me''); each item in this list then becomes a conversational entity for subsequent turns.
     \item Background Entities: These are relevant entities that come from background processes that might not necessarily be a direct part of what the user sees on their screen or their interaction with the virtual agent; for example, an alarm that starts ringing or music that is playing in the background.
 \end{enumerate}
 
We pose the task of reference resolution as a multiple choice task for the LLM, where the intended output is a single option (or multiple options) from the entities shown on the user's screen. In some cases, the answer could also be "None of these", in which case the model needs to predict ``0''. 

To evaluate this task, we check if the predicted set of options matches the ground truth set; in other words, we allow the model to output the relevant entities in any order, i.e. if the Ground Truth is entities 8, 7, and 4, then we accept any permutation of these three correct entities while evaluating the performance of the model. 

Note that as in \citet{ates2023marrs,bhargava2023referring}, we assume that entities along with their types come in from an upstream system (for example, through a mechanism involving entity pullers which are able to extract entities in a high recall manner or through a donation from a device app, as in \citet{aas2023intelligent}).

\section{Datasets} \label{sec:datasets}

Our datasets comprise data that was either synthetically created, or created with the help of annotators. Each data point contains the user query and a list of entities, along with the ground-truth entity (or set of entities) that are relevant to the corresponding user query. Each entity, in turn, contains information about its type and other properties such as the name and other textual details associated with the entity (the label and time of an alarm, for example). For data points where relevant on-screen context exists, this context is available in the form of the bounding box of the entity, and the list of objects surrounding it along with properties of these surrounding objects such as their types, textual contents and locations. Note that our data collection follows that of \citet{bhargava2023referring,ates2023marrs}; we present an overview here and direct the interested reader to the aforementioned papers for a more detailed description. Note also that each dataset below is somewhat representative of one of our tasks of interest (with our synthetic data bucket being used for both conversational and background entity resolution).

\begin{table}[!thb]
    \centering
    \caption{Dataset Sizes (Train Set and Test Set)}
    \label{tab:datasetsizes}
    \begin{tabular}{lcc}
        \toprule
        Dataset & Train & Test \\
        \midrule
        Conversational & 2.3k & 1.2k \\
        Synthetic & 3.9k & 1.1k \\
        On-screen & 10.1k & 1.9k \\
        \bottomrule
    \end{tabular}
\end{table}

\subsection{Conversational Data} 

In this case, data is collected for entities that are relevant to the user interaction with the agent. To do this, annotators are shown sample conversations between a user and an agent with synthetic lists of entities provided, and asked to provide queries that unambiguously reference an arbitrarily picked entity in the aforementioned synthetic list. Annotators might thus be provided with a synthesized list of businesses or alarms and asked to refer to a particular entity within that list.

For example, the annotator might be shown a list of businesses that are synthetically constructed, and then asked to refer to a specific one in the list provided; for instance, they might say ``Take me to the one that's second from the bottom'' or ``Call the one on Main Street''.

\subsection{Synthetic Data}

Another approach to obtain data is to rely on synthetic data from templates. This approach is particularly useful for type-based references, when the user query and the entity type are sufficient to resolve the reference, and descriptions are not relied upon. Note that the synthetic nature of this dataset does not preclude it from containing datapoints in which multiple entities can be resolved to a given reference: for example, for the query ``play it'', ``it'' can be resolved to all entities of both the types ``music'' and ``video''.

The pipeline used to generate the synthetic data comprises of two parts: a set of templates and a list accompanying each template. The first part, a ``language template'', contains different variations of queries that can be used for targeted cases, with slots present that can be filled pragmatically from those defined in the ``slot list''. The second, a ``slot list'' accompanying the aforementioned template, includes mentions and other possible slot values (often comprising of named entities that aren't mentions, or other slots that can take a large number of possible values such as date-times) if necessary. The slot list also contains the ground truth entity (or entities) that the mentions listed, when filled into the language template, could resolve to.

The data generation pipeline then takes the language template and slot list, and uses them to generate the possible queries. It does this by substituting corresponding values from the slot lists into the language templates to obtain fully formed user queries. The corresponding synthetic data is formed by using these queries and the ground truth entities present in the slot list, and adding in entities of other types into the data to serve as random negatives.

For example, a given language template might consist of phrases like ``share [mention] with [name]'' and ``send [mention] to [name] please''. The corresponding slot list might have ``[mention]'' mapping to ``this address'' and ``that address'', ``[name]'' mapping to various person names, and the ground truth entity tagged as ``email address'' and ``physical address''. The pipeline then generates queries like ``share that address with Mom'' with ``email address'' and ``physical address'' entities marked as possible ground truth entity types, and entities of other types marked as negative.

\begin{figure}[!htb]
\centering
\begin{subfigure}{.23\textwidth}
  \centering
  \includegraphics[width=0.98\textwidth]{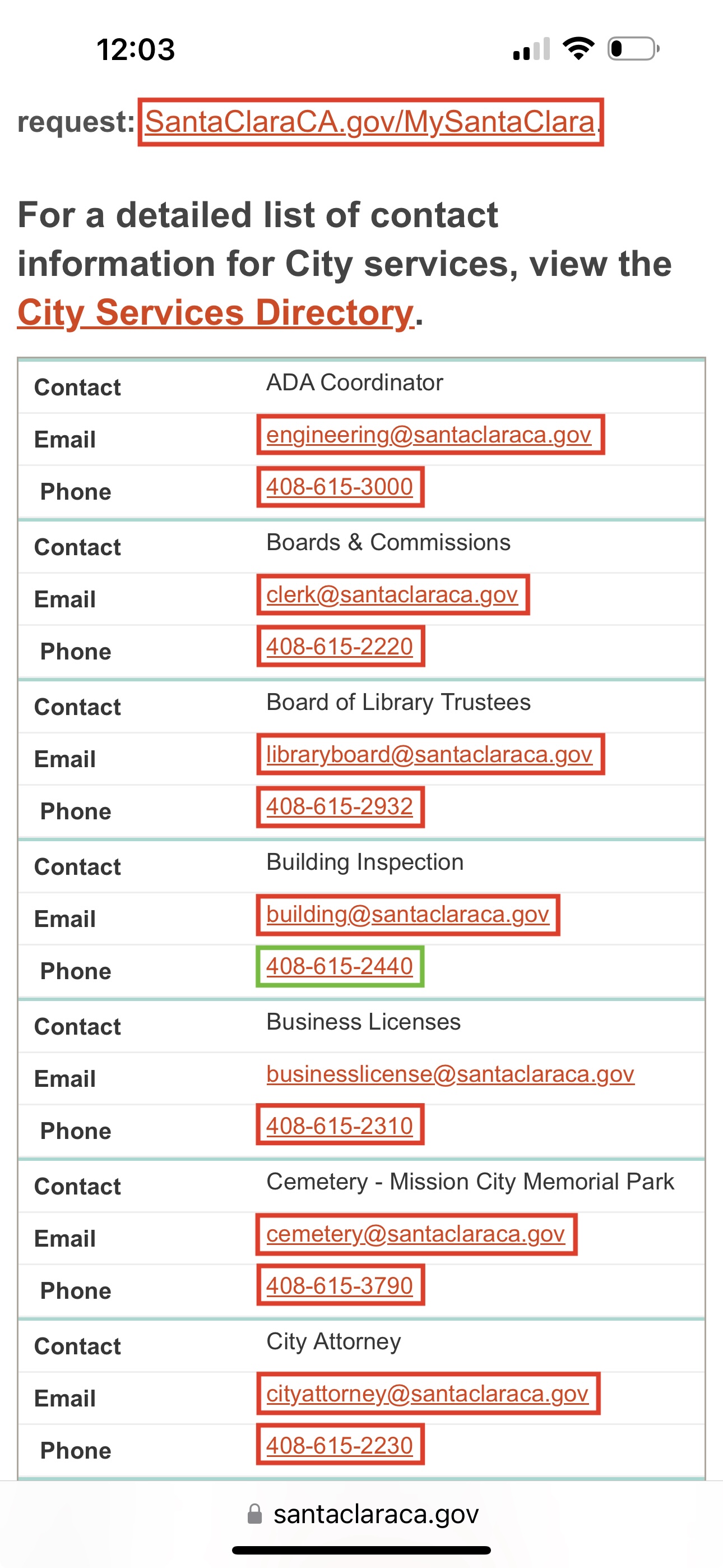}
  \caption{Screenshot example used in first annotation project}
  \label{fig:screenshot_annotation}
\end{subfigure}%
\hfill
\begin{subfigure}{.23\textwidth}
  \centering
  \includegraphics[width=0.98\textwidth]{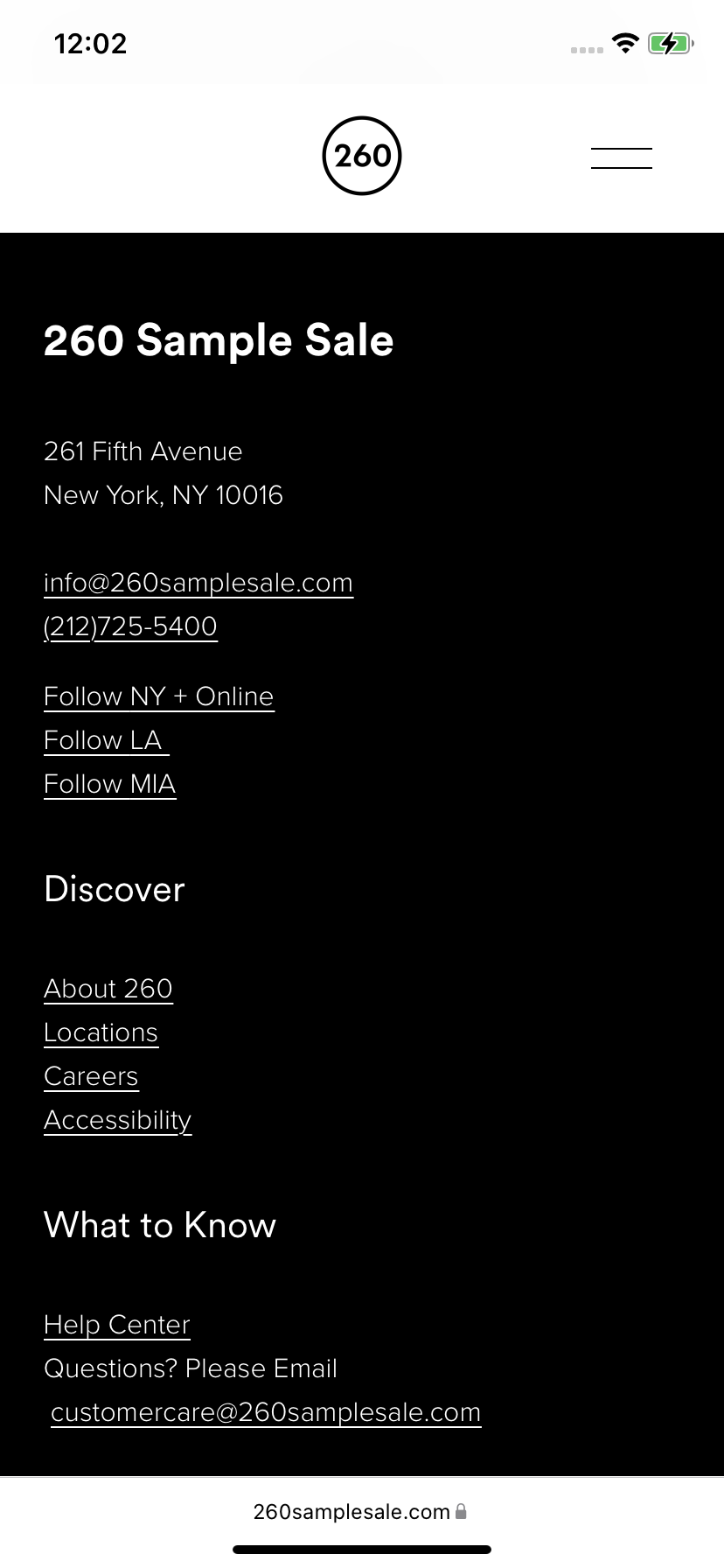}
  \caption{Screenshot example used in second annotation project}
  \label{fig:screenshot_annotation_step_2}
\end{subfigure}
\caption{Sample screenshots used in the annotation of on-screen data. The data was annotated in a two-step process, as described in Section~\ref{sec:onscreen_data}.}
\label{fig:test}
\end{figure}

\subsection{On-screen Data} \label{sec:onscreen_data}

As in \citet{bhargava2023referring}, screen data were collected from various web pages where phone number, e-mail and/or physical address information exist. Our on-screen data annotation comprised of a two-phase process. The first phase was used to obtain queries based on the screens shown, and the second one was for identifying the entities and mention for the given query. In the first grading project, annotators were given a screenshot (Figure~\ref{fig:screenshot_annotation}) with green and red boxes, and were asked to classify the green boxed data into one of the entities such as phone number, email address, etc. Then, annotators were then asked to provide three unique queries for the green boxed data. 

In the second annotation project (Figure~\ref{fig:screenshot_annotation_step_2}), queries collected in the first step were shown to annotators one by one with their corresponding screenshots (but this time, without the bounding boxes), and with all the screen entities as a list. The annotators were asked if the query contains a mention to one of the given visual entities, and if the query sound natural. They were also asked to provide the entities from the list that were referred to in the given query, and to tag the part of the query referring that entity.

\section{Models}

We compare our proposed model ReALM, described in detail in Section~\ref{sec:realm} below, with two baseline approaches: one based on the reference resolver proposed in MARRS (Section~\ref{sec:marrs}), which is non-LLM based, and one based on 
ChatGPT (both GPT-3.5 and GPT-4; Section~\ref{sec:gpt}).

\begin{figure}[!htb]
    \centering
    \begin{subfigure}[b]{0.23\textwidth}
        \includegraphics[width=\textwidth]{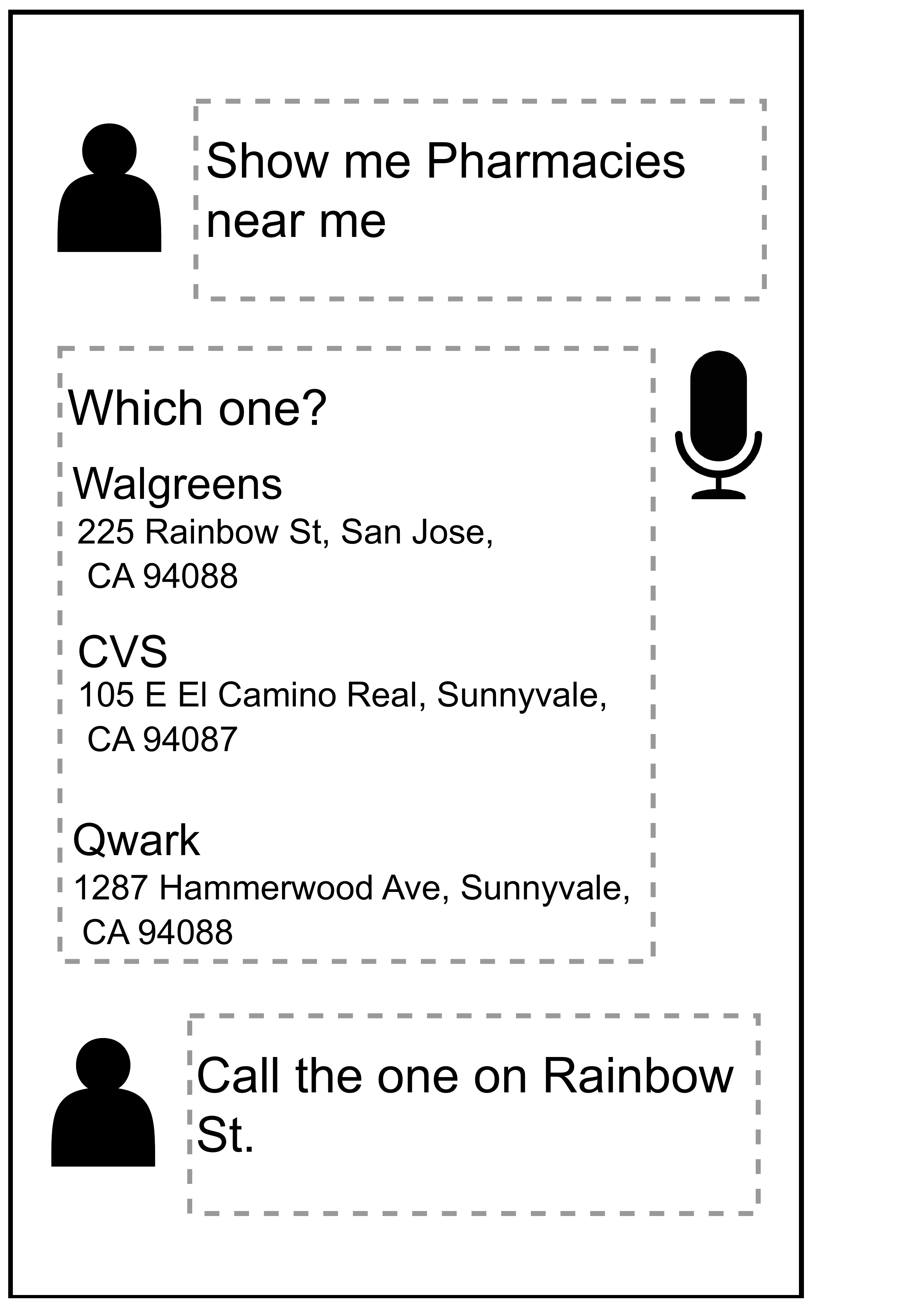}
        \caption{Conversational User Turns}
    \end{subfigure}
    \hfill
    \begin{subfigure}[b]{0.20\textwidth}
        \includegraphics[width=\textwidth]{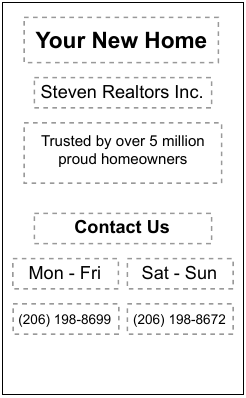}
        \caption{Onscreen Capture}
    \end{subfigure}

    \caption{Technical diagrams representing user turns with a conversational assistant in (a), and a user screen in (b). Shaded rectangles represent various elements shown on the screen detectable by screen parser-extractors.}
    \label{fig:dataset_examples}
\end{figure}

\subsection{MARRS} \label{sec:marrs}

As a baseline, we compare against the system proposed in \citet{ates2023marrs}, in turn a variation of \citet{bhargava2023referring}, both of which are non-LLM based approaches. While the latter approach focuses on on-screen entities, MARRS extends it to conversational and background entities as well. For our baseline comparison, we trained a re-implementation of this system with the datasets described in Section~\ref{sec:datasets}, which includes conversation, on-screen and synthetic data. Note that in contrast to our approach, which uses a generic off-the-shelf LLM, this baseline we compare against was specifically designed for the task of reference resolution.

\subsection{ChatGPT} \label{sec:gpt}

As another baseline, we run the GPT-3.5 \cite{brown2020language,ouyang2022training} and GPT-4 \cite{achiam2023gpt} variants of ChatGPT, as available on \gptdate, with in-context learning. As in our setup, we aim to get both variants to predict a list of entities from a set that is available. In the case of GPT-3.5, which only accepts text, our input consists of the prompt alone; however, in the case of GPT-4, which also has the ability to contextualize on images, we provide the system with a screenshot for the task of on-screen reference resolution, which we find helps substantially improve performance. Note that our ChatGPT prompt and prompt+image formulation are, to the best of our knowledge, in and of themselves novel. While we believe it might be possible to further improve results, for example, by sampling semantically similar queries up until we hit the prompt length, this more complex approach deserves further, dedicated exploration, and we leave this to future work. 

\subsection{Our Approach} \label{sec:realm}

In this section, we provide examples of conversational and onscreen reference resolution tasks, followed by how we prompt the model to resolve the same.

We use the following pipeline for fine-tuning an LLM (a FLAN-T5 model \cite{chung2022scaling}) in our case. We provide the parsed input to our model, and finetune it. Note that unlike for the baseline, we do not run an extensive hyperparameter search on the FLAN-T5 model, sticking to the default fine-tuning parameters.
\\
\\
{
\small
\noindent \fbox{\begin{minipage}{0.46\textwidth}

Select which among the following entities, if any, are required to understand the user request below. Output 0 if none of the entities are relevant. 

 User request: Call the one on Rainbow St

 User Entities:

 0. None

 1. Type: Local Business | Name: Walgreens | Address: 225 Rainbow St, San Jose CA 94088

 2. Type: Local Business | Name: CVS | Address: 105 E El Camino Real, Sunnyvale, CA 94087

 3. Type: Local Business | Name: Qwark | Address: 1287 Hammerwood Ave, Sunnyvale, CA

 Relevant entity: 
 
\end{minipage}}
 }
\\
\\
\\
 {
\small
\noindent \fbox{\begin{minipage}{0.46\textwidth}

Select which among the following entities, if any, are required to understand the user request below. Output 0 if none of the entities are relevant. 

 User request:  Save the phone number at the bottom-right

 Screen: \\
         Your New home! \\
        Steven Realtors Inc. \\
        Trusted by over 5 million \\
        Proud homeowners \\
        Contact Us \\
        Monday - \hspace{8em} Saturday - \\
        Friday \hspace{9.3em} Sunday \\
        \{\{1. (206) 198 1999\}\} \hspace{2em} \{\{2. (206) 198 1699\}\} 

 Relevant entity: 
 
\end{minipage}}
 }
 \\
 \\

Each data point consisting of a user query and the corresponding entities is converted into a sentence-wise format that we can feed to an LLM for training. Examples of the input before and after processing are shown in Appendix Sections \ref{sec:appendix_other_strategies} and \ref{sec:appendix_sample_inputs}, with examples of how we convert entities of different types into text shown in Appendix~\ref{sec:app_entity_reps}. Note that the entities are shuffled before being sent to the model so that the model does not overfit to particular entity positions.

With respect to the output that the model predicts, empirically, we find that the model is consistently able to predict a valid integer (or list of integers), without deviating and outputting any other text. In addition, we observe that the model also respects general output constraints (such as not predicting a `0` that represents `None of These` at the same time as one or more other entities) as well as those constrains enforced by the input (such as ensuring all predicted entity indices actually exist on the input side). The one exception that we observe is that, on occasion, we find that the model predicts the same entity twice (successively) in its output list. The only post-processing heuristic we apply is thus to convert the model's predictions into a set of unique entities.

\subsubsection{Conversational References}

For the sake of this work, we assume conversational references to be of two types: type-based and descriptive. Type-based references are heavily reliant on using the user query in conjunction with the types of the entities to identify which entity (of a set of entities) are most relevant to the user query in question: for example, if the user says ``play this'', we know that they are referring to an entity like a song or a movie, as opposed to a phone number or an address; ``call him'' likewise refers to a  contact or possibly a phone number, as opposed to an alarm. Descriptive references, in contrast, tend to use a property of the entity to uniquely identify it: ``The one in Times Square'' for example might help uniquely refer to one among a set of addresses or business. Note that it is often the case that references might rely on both types and descriptions to unambiguously refer to a single object: consider the examples ``play the one from Abbey Road'' vs ``directions to the one on Abbey Road'', both of which rely on both the entity type and description to identify a song in the first case and address in the second. In our proposed approach, we simply encode the type and various properties of the entity. We show our detailed encoding scheme in Appendix~\ref{sec:app_entity_reps}.

\subsubsection{Onscreen References} \label{sec:onscreen}

For onscreen references, as in \citet{bhargava2023referring}, we assume the presence of upstream data detectors that are able to parse screen text to extract entities. These entities are then available along with their types, bounding boxes and a list of non-entity text elements surrounding the entity in question.

To encode these entities (and thereby, the relevant parts of the screen) into the LM in a manner that involves text alone, we use the novel algorithm given in Algorithm~\ref{algo:onscreen}. Intuitively, we assume the location of all entities and their surrounding objects to be representible by the center of their respective bounding boxes. We then sort these centers (and thereby, the associated objects) from top-to-bottom (i.e., vertically, along the y-axis), and the use a stable sort to sort from left-to-right (i.e., horizontally, along the x-axis). Next, all objects that are within a \texttt{margin} are treated as being on the same line, and are separated from each other by a tab; objects further down outside the margin are placed on the next line, and this is repeatedly, effectively encoding the screen in a left-to-right, top-to-bottom fashion in plain text.

\begin{algorithm}[!]
  \caption{Onscreen Parse Construction with Turn Object Injection} \label{algo:onscreen}
  \KwData{List of turn objects}
  \KwResult{Onscreen parse}
  $onscreen\_parse \gets$ Empty list of onscreen parse elements\;

  \tcp{Step 0: Get all text boxes present in the screen}
  \For{each turn object $t$, index $i$}{
    \tcp{Step 1: Get unique surrounding objects}
    $surrounding\_objects \gets$ Set of surrounding objects for $t$\;
    
    \tcp{Step 2: Insert turn objects into the set}
    $surrounding\_objects \gets surrounding\_objects \cup \{[[i. t]]\}$\;
    }
    
    \tcp{Step 3: Sorting the centers of all surrounding objects}
    $sorted\_objects \gets$ Sort objects in $surrounding\_objects$ by center (Top $\rightarrow$ Bottom, Left $\rightarrow$ Right)\;
    
    \tcp{Step 4: Determine vertical levels}
    $margin \gets$ Margin for considering objects at the same level\;
    $levels \gets$ List of vertical levels\;
    \For{each object $o$ in $sorted\_objects$}{
      $same\_level \gets$ List of objects at the same level as $o$\;
      \For{each object $other$ in $sorted\_objects$}{
        \If{$o$ is not the same as $other$ and $|o.center\_top - other.center\_top| \leq margin$}{
          $same\_level \gets same\_level \cup \{other\}$\;
        }
      }
      $levels \gets levels \cup \{same\_level\}$\;
    }
    
    \tcp{Step 5: Construct onscreen parse}
    \For{each level $l$ in $levels$}{
      $level\_parse \gets$ Empty string\;
      \For{each object $obj$ in $l$}{
        $level\_parse \gets level\_parse + \text{"\textbackslash t"} + obj$\;
      }
      $onscreen\_parse \gets onscreen\_parse +  \text{"\textbackslash n"} + level\_parse$\;
    }
  \KwRet $onscreen\_parse$\;
\end{algorithm}

\section{Results}


\begin{table}[H]
    \centering
    \caption{Model Accuracy for Different Datasets. A prediction is correct if the model correctly predicts all relevant entities, and incorrect otherwise. \textbf{Conv} refers to the Conversational Dataset, \textbf{Synth} to the Synthetic one, \textbf{Screen} to the Onscreen one and \textbf{Unseen} to a conversational dataset pertaining to a held-out domain.}
    \label{tab:modelaccuracy}
    \addtolength{\tabcolsep}{-0.2em}
    \begin{tabular}{l@{\hskip1pt}cccc@{ }}
        \toprule
        \textbf{Model} & \textbf{Conv} & \textbf{Synth} & \textbf{Screen} & \textbf{Unseen} \\
        \midrule
        \textbf{MARRS} & 92.1 & 99.4 & 83.5 & 84.5 \\
        \hdashline
        \textbf{GPT-3.5} & 84.1 & 34.2 & 74.1 & 67.5 \\
        \textbf{GPT-4} & 97.0 & 58.7 & 90.1 & 98.4 \\
        \hdashline
        \textbf{ReALM-80M} & 96.7 & 99.5 & 88.9 & 99.3 \\
        \textbf{ReALM-250M} & 97.8 & 99.8 & 90.6 & 97.2 \\
        \textbf{ReALM-1B} & 97.9 & 99.7 & 91.4 & 94.8 \\
        \textbf{ReALM-3B} & 97.9 & 99.8 & 93.0 & 97.8 \\
        \bottomrule
    \end{tabular}
\end{table}

    \begin{table*}[!t]
  \centering
  \begin{subtable}{0.48\textwidth}
    \centering
    \small
    \caption{Semantic Understanding}
    \label{tab:userrequestentities3a}
    \begin{tabular}{l}
        \toprule
        \textbf{User Request:} Call the evening Number \\
        \midrule
        \textbf{Screen:} \\
        \{\{1. 9 AM - 5 PM\}\} \\
        \{\{2. 901.969.3120\}\} \\
        \{\{3. 5 PM - 9 PM\}\} \\
        \{\{4. 901.969.3391\}\} \\
        \midrule
        \textbf{Model Output:} 4 \\
        \bottomrule
    \end{tabular}
  \end{subtable}
  \hfill
  \begin{subtable}{0.48\textwidth}
    \centering
    \small
    \caption{Summarisation}
    \label{tab:userrequestentities3b}
    \begin{tabular}{l}
        \toprule
        \textbf{User Request:} Remind me to get printouts \\ before the tax deadline \\
        \midrule
        \textbf{Screen:} \\
        Tax Deadlines 2023 \\
        \{\{1. Feb 15\}\} \\
        Reclaim your tax exemption from withholding \\
        \{\{2. April 18\}\} \\
        First-quarter estimated tax payment due \\
        \midrule
        \textbf{Model Output:} 2 \\
        \bottomrule
    \end{tabular}
  \end{subtable}

  \begin{subtable}{0.48\textwidth}
    \centering
    \small
    \caption{World Understanding}
    \label{tab:userrequestentities3c}
    \begin{tabular}{l}
        \toprule
        \textbf{User Request:} Take me to the one in Washington \\
        \midrule
        \textbf{Screen:} \\
        Indian Embassy \\
        \{\{1. 1701 El Camino Real, Mountain View 94040\}\} \\
        \{\{2. 333 Dexter Ave N, Seattle 98109\}\} \\
        \{\{3. 8295 Tournament Drive, Memphis, TN 38125\}\} \\
        \midrule
        \textbf{Model Output:} 2 \\
        \bottomrule
    \end{tabular}
  \end{subtable}
  \hfill
  \begin{subtable}{0.48\textwidth}
    \centering
    \small
    \caption{Commonsense Reasoning}
    \label{tab:userrequestentities3d}
    \begin{tabular}{l}
        \toprule
        \textbf{User Request:} Save the link to the breakfast Recipe \\
        \midrule
        \textbf{Screen:} \\
        IMAGE \\
        Strawberry Granola \\
        \{\{1. Recipe link\}\} \\
        IMAGE \\
        Lavender boba tea \\
        \{\{2. Recipe link\}\} \\
        \midrule
        \textbf{Model Output:} 1 \\
        \bottomrule
    \end{tabular}
  \end{subtable}

  \caption{Qualitative examples that demonstrate the ability of ReALM to adapt to complex use-cases.}
  \label{sec:appendix_new_experiences}
\end{table*}

We present our results in Table~\ref{tab:modelaccuracy}. 
Overall, we find that our approach outperforms the MARRS model in all types of datasets. We also find that our approach is able to outperform GPT-3.5, which has a significantly larger number of parameters than our model by several orders of magnitude. We also find that our approach performs in the same ballpark as the latest GPT-4 despite being a much lighter (and faster) model. We especially wish to highlight the gains on onscreen datasets, and find that our model with the textual encoding approach is able to perform almost as well as GPT-4 despite the latter being provided with screenshots. \\
Additionally, we also experiment with models of different sizes. 
We see that while performance in general improves across all dataset families with an increase in model size, the difference is most pronounced for the onscreen datasets, which alludes to the task being more complex in nature. Interestingly, and contrary to an otherwise consistent trend of larger models performing better, we find that performance on our Unseen dataset, which contains a held-out domain, first decreases with an increase in model size before increasing again. We hypothesize that this is due to the double-descent phenomenon \cite{nakkiran2019deep}.

\subsection{Analysis}

    \textbf{GPT-4 $\approx$ ReALM $\gg$ MARRS for new use-cases: }
    As a case study, we explore zero-shot performance of this model on an unseen domain: Alarms (we show a sample data point in Appendix Table \ref{tab:alarm_example}). The last column in Table \ref{tab:modelaccuracy} compares the performance of all approaches and baselines on this unseen test set. We find that all of the LLM-based approaches outperform the FT model for this test set. Among the two, we find that the performance of ReALM and GPT-4 are very similar for the unseen domain. Additionally, Table~\ref{sec:appendix_new_experiences} shows completely new experiences enabled by ReALM due to the LLM's superior ability to perform complex understanding of natural language. 

    \textbf{ReALM $>$ GPT-4 for domain-specific queries} We find that due to finetuning on user requests, ReALM is able to understand more domain-specific questions. Consider Table~\ref{tab:qual_settings}. GPT-4 incorrectly assumes the reference to be about only a setting, whereas the ground truth consists of a home automation device in the background as well, and GPT-4 lacks the domain knowledge to be able to recognise that the device would also be relevant to this reference. ReALM, in contrast, doesn't suffer from this due to being trained on domain-specific data.

\begin{table}[!t]
    \small
    \caption{User Request for Setting or Home Device}
    \label{tab:qual_settings}
    \begin{tabular}{l}
        \toprule
        \textbf{User Request:} Can you make it brighter? \\
        \midrule
        \textbf{Entities Shown to User:}  \\
        1. Type: Settings \\
        2. Type: UserEntity | homeAutomationAccessoryName \\
        \midrule
        \textbf{GPT-4 Prediction: 1}
        \textbf{Ground Truth: } 1, 2 \\
        \bottomrule
    \end{tabular} \\
\end{table}

\section{Conclusion and Future Work}

In this work, we demonstrate how large language models, which are typically trained on text alone, can be also be adapted to perform reference resolution to items in an extra-linguistic context. We do this by encoding entity candidates as natural text; we demonstrate how entities that are present on the screen can be passed into an LLM using a novel textual representation that effectively summarizes the user's screen while retaining relative spatial positions of these entities. Our proposed system is thus able to resolve references in a variety of human-computer interaction settings, such as those involving on-screen, conversational and background entities; we note, however, that our proposed approach focuses primarily on anaphoric and deictic references, and we leave the extension of our system to handle other types of references, such as bridging references, to future work.

In addition, we show that ReALM outperforms previous approaches, and performs roughly as well as the current state-of-the-art LLM, GPT-4, despite consisting of far fewer parameters, even for onscreen references despite being purely in the textual domain. It also outperforms GPT-4 for domain-specific user queries, thus making ReALM an ideal choice for a practical reference resolution system that can exist on-device without compromising on performance.

While our approach is effective in encoding the position of entities on the screen, we find that it may not be able to resolve complex user queries that rely on nuanced positional understanding. We thus believe that exploring more complex approaches such as splitting the screen into a grid and encoding these relative spatial positions into text, while challenging, is a promising avenue of future exploration. In addition, in contrast to a critical assumption of our proposed system, not all on-screen entities are textual. While extending this paper to cover on-screen images, graphics and UI elements is beyond the scope of this work, this is certainly another extension that merits further investigation. 

\section*{Ethics Statement}

While LLMs can generate unexpected output including potentially harmful text, our system offers the ability to constrain decoding or use simple post-processing to ensure this does not happen. Note however that practically we find very little hallucination or even text that deviates from the format that the models were finetuned on, and thus do not constrain the decoding of the LLM.

\section*{Acknowledgements}

The authors would like to thank Nidhi Rajshree, Stephen Pulman, Leon Liyang Zhang, Jiarui Lu, Jeff Nichols, Shruti Bhargava, Dhivya Piraviperumal, Junhan Chen and the anonymous reviewers for their help, suggestions, and feedback.

\bibliography{anthology,custom}

\appendix

\include{paper_appendix}

\end{document}

%% file: paper_appendix.tex
\section{Encoding onscreen entities} 
\label{sec:appendix_other_strategies}

\begin{figure*}[!htb]
    \centering
    
    \begin{subfigure}[b]{0.4\textwidth}
        \includegraphics[width=\textwidth]{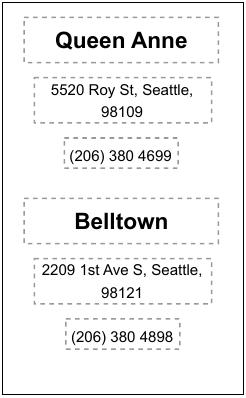}
        \caption{Onscreen Capture 1}
    \end{subfigure}
    \hfill
    \begin{subfigure}[b]{0.4\textwidth}
        \includegraphics[width=\textwidth]{figures/onscreen_example.pdf}
        \caption{Onscreen Capture 2}
    \end{subfigure}

    \caption{Technical diagrams representing user screens. Shaded rectangles represent various elements shown on the screen detectable by screen parser-extractors.}
    \label{fig:onscreen_example}
\end{figure*}

First, we show sample representations of what a screen grab might look like, as parsed and exposed to the system. We show these representations in Figure~\ref{fig:onscreen_example}

We now describe some other strategies of encoding on-screen elements that we explored.

\begin{itemize}
    \item \textbf{Clustering: } We explored a clustering-based approach wherein we perforedm a spatial clustering of the various surrounding objects present in the screen. We did this to establish semantic clusters wherein a user could refer to nearby bounding boxes (such as the contact information) by a particular title. The detailed approach is given in Algorithm~\ref{algo:dbscan}, and a sample encoding is shown in Table \ref{tab:clustering}. The biggest drawback of the approach was that the prompt length often explodes as the number of entities in a cluster increases, as each of the objects in the cluster would have every other object in its surrounding objects. 

    \item \textbf{Onscreen Grab: } To mitigate this issue, we employed a second approach (similar to our final approach), wherein we parsed the screen as in our final approach, the only difference being that we didn't annotate the turn objects within the parse itself, but provided the turn objects as a list instead (see Table \ref{tab:onscreen_grab}). 

    \item \textbf{Onscreen Grab with Injected Turn Objects: }Finally, the exact algorithm employed in our final approach is given in \ref{algo:onscreen}, and a sample encoding is shown in Table \ref{tab:injected_onscreen}. 
\end{itemize}

\begin{table}[!b]
\small
    \centering
    \caption{Clustering-based encoding}
    \label{tab:clustering}
    \begin{tabular}{l}
        \toprule
        \textbf{User Request:} Get me directions to the branch in \\ Queen Anne \\
        \midrule
        \textbf{Entities Shown to User:}  \\
        1. Type: Postal Address | Value: 5520 Roy St, \\ Seattle 98109 |
        surr\_objects: Queen Anne, (206) 380 4699 \\
        2. Type: Phone Number | Value: (206) 380 4699 \\
        surr\_objects: Queen Anne, 5520 Roy St, Seattle 98109 \\
        3. Type: Phone Number | Value: (206) 380 4898 \\
        surr\_objects: Belltown, 2209 1st Ave S, Seattle 98121 \\
        4. Type: Postal Address | Value: 2209 1st Ave, \\ Seattle 98121 |
        surr\_objects: Belltown, (206) 380 4898 \\
        \midrule
        \textbf{Ground Truth: } 1 \\
        \bottomrule
    \end{tabular}
\end{table}

\begin{table}[!h]
\small
    \centering
    \caption{Onscreen Grab encoding}
    \label{tab:onscreen_grab}
    \begin{tabular}{l}
        \toprule
        \textbf{User Request:} Save the phone number at the bottom-right \\
        \midrule
        \textbf{Screen:} \\
        Your New home! \\
        Steven Realtors Inc. \\
        Trusted by over 5 million \\
        Proud homeowners \\
        Contact Us \\
        Monday - \hspace{3.5em} Saturday - \\
        Friday \hspace{4.8em} Sunday \\
        (206) 198 1699 \hspace{1.5em} (206) 198 1999 \\
        \midrule
        \textbf{Entities Shown to User:}  \\
        1. Type: Phone Number | Value: (206) 198 1999 \\
        2. Type: Phone Number | Value: (206) 198 1699 \\
        \midrule
        \textbf{Ground Truth: } 1, 2 \\
        \bottomrule
    \end{tabular}
\end{table}

\begin{table}[!b]
\small
    \centering
    \caption{Injected Onscreen Encoding (Final Approach)}
    \label{tab:injected_onscreen}
    \begin{tabular}{l}
        \toprule
        \textbf{User Request:} Save the phone number at the bottom-right \\
        \midrule
        \textbf{Screen:} \\
        Your New home! \\
        Steven Realtors Inc. \\
        Trusted by over 5 million \\
        Proud homeowners \\
        Contact Us \\
        Monday - \hspace{8em} Saturday - \\
        Friday \hspace{9.3em} Sunday \\
        \{\{1. (206) 198 1999\}\} \hspace{2em} \{\{2. (206) 198 1699\}\} \\
        \midrule
        \textbf{Ground Truth: } 1, 2 \\
        \bottomrule
    \end{tabular}
\end{table}

\begin{algorithm}[!h]
  \caption{Surrounding Object Clustering and Prompt Generation}  \label{algo:dbscan}
  \KwData{List of MDF turn objects}
  \KwResult{Updated turn objects with surrounding object prompts}
  \For{each MDF turn object $t$}{
    \tcp{Step 1: Get unique surrounding objects}
    $surrounding\_objects \gets$ Set of unique surrounding objects for $t$\;
    
    \tcp{Step 2: Spatially cluster surrounding object bounding boxes}
    $clusters \gets DBScan(surrounding\_objects, $ \\ $rect\_distance)$\;
    
    \tcp{Step 3: Predict the cluster for turn object}
    $t\_cluster \gets$ Predicted cluster for $t$\;
    
    \For{each surrounding object $s$ in $surrounding\_objects$}{
      \If{$s$ belongs to cluster $t\_cluster$}{
        \tcp{Step 4: Process non-overlapping surrounding objects}
        \If{no string overlap between $t$ and $s$}{
          Add $s$ to the prompt under key `surrounding\_object`\;
        }
      }
    }
    
    \tcp{Step 5: Provide global positioning information}
    $t.distance\_from\_top \gets$ Compute distance from the top for $t$\;
    $t.distance\_from\_left \gets$ Compute distance from the left for $t$\;
  }
    \KwRet $prompt$\;
\end{algorithm}

\begin{figure}[!h]
\centering
  \includegraphics[width=0.48\textwidth]{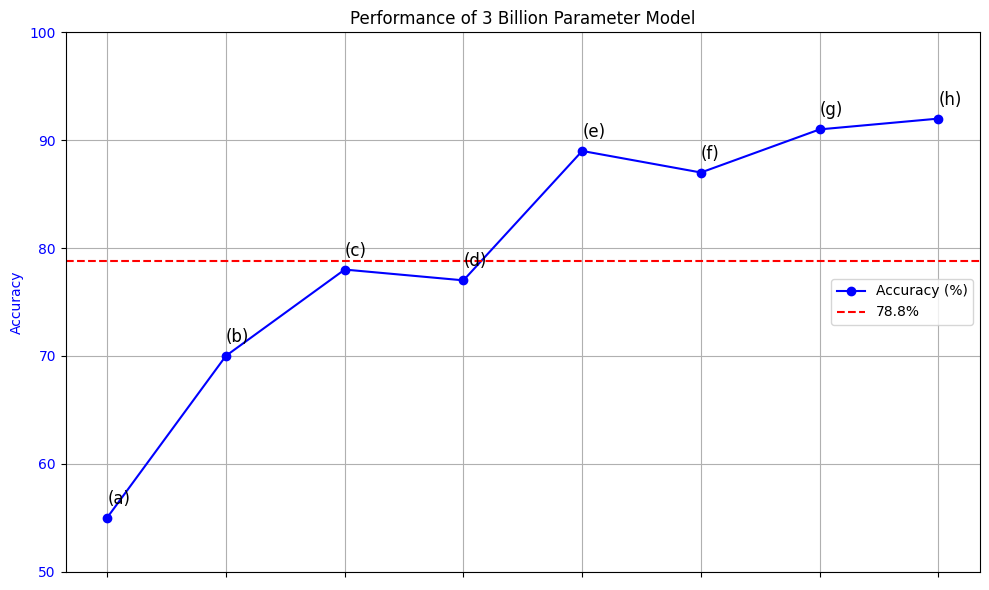}
\caption{Performance improvements with each experiment -- (a) Baseline Finetuned LLM, (b) Obtaining screen elements through OCR, 
(c) Obtaining screen elements through UI elements and Clustering (d) Adding an extra newline between the instruction and user request, (e) Onscreen Grab, (f) Onscreen Grab with injected turn objects, (g) Onscreen Grab with injected turn object + needing lines to be separated by at least Margin, (h) Separating elements in the same line by a tab}
  \label{fig:ablation}
\end{figure}

We show an ablation in Figure~\ref{fig:ablation}, in which we show the performance of the various encoding approaches described above (and some other hill-climbing efforts). 

We show the algorithm used to encode onscreen entities, described in Section~\ref{sec:onscreen}, in Algorithm~\ref{algo:onscreen}.

\clearpage
\newpage

\section{Entity Representations} \label{sec:app_entity_reps}

In Table~\ref{tab:entitytype}, we show some examples of various domains and their representations, as fed into the LLM.

\begin{table*}[!htb]
\small
\centering
\caption{Entity Domains and their Representations}
\label{tab:entitytype}
\begin{tabular}{ll}
\toprule
Entity Type & After \\
\midrule
alarm & Type: Alarm | time: 08:06 PM; label: brush hair; status: Off \\
app & Type: App | clock \\
book & Type: Book \\
date time & Type: DateTime | 1 | 1 | 2021 \\
email address & Type: EmailAddress | membership@ipsa.org \\
flight number & Type: FlightNumber \\
general text & Type: GeneralText \\
home device & Type: UserEntity | heater \\
home room & Type: UserEntity | Db Bedroom \\
local business & Type: LocalBusiness | PostalAddress: 15 Broad St, Albany 31701 | Ameris Bank | list\_position: 13 \\
media album & Type: MediaItem | MediaItemType: MediaItemType\_Album | Mellon Collie \\
package & Type: Package \\
painting & Type: Painting \\
person & Type: Person | Sebastian \\
phone number & Type: PhoneNumber | 955 545 060 \\
photo & Type: Photo \\
physical address & Type: PostalAddress | GeographicArea: 814 Elmwood Ave, NY, 14222 \\
plant animal & Type: PlantAnimal \\
setting & Type: Setting | dark mode \\
tracking number & Type: TrackingNumber \\
url & Type: Uri | NY.gov \\
\bottomrule
\end{tabular}
\end{table*}

\clearpage
\newpage

\section{Sample Inputs} \label{sec:appendix_sample_inputs}

In this section, we show examples of how inputs into the model have been encoded, in the form of a visual representation in Tables~\ref{tab:single_gt_input}, \ref{tab:multi_gt_input} and \ref{tab:alarm_example}.

 \begin{table*}[!ht]
    \centering
    \caption{Sample input with single ground truth}
    \label{tab:single_gt_input}
    \begin{tabular}{l}
        \toprule
        \textbf{User Request:} Call the one on Rainbow St. \\
        \midrule
        \textbf{Entities Shown to User:}  \\
        1. Type: Local Business | Name: Walgreens | Address: 225 Rainbow St, San Jose CA 94088 \\
        2. Type: Local Business | Name: CVS | Address: 105 E El Camino Real, Sunnyvale, CA 94087 \\
        3. Type: Local Business | Name: Qwark | Address: 1287 Hammerwood Ave, Sunnyvale, CA 94089 \\
        \midrule
        \textbf{Ground Truth: } 1 \\
        \bottomrule
    \end{tabular}
\end{table*}

\begin{table*}[!ht]
    \caption{Sample input with multiple ground truths}
    \label{tab:multi_gt_input}
    \begin{tabular}{l}
        \toprule
        \textbf{User Request:} Save the address. \\
        \midrule
        \textbf{Entities Shown to User:}  \\
        1. Type: Postal Address | Value: 225 Rainbow St, San Jose CA 94088 \\
        2. Type: Email Address | Value: \texttt{contactus@cvs.com} \\
        3. Type: URL | Value: \texttt{cvspharmacies.com/usa} \\
        \midrule
        \textbf{Ground Truth: } 1, 2, 3 \\
        \bottomrule
    \end{tabular}
\end{table*}

\begin{table}[!h]
    \small
    \caption{User Request for Alarms}
    \label{tab:alarm_example}
    \begin{tabular}{l}
        \toprule
        \textbf{User Request:} Switch off the one reminding me to \\ pick up didi. \\
        \midrule
        \textbf{Entities Shown to User:}  \\
        1. Type: Alarm | open laptop \\
        2. Type: Alarm | text Lauren to shower \\
        3. Type: Alarm | pick up didi \\
        4. Type: Alarm | forget this \\
        \midrule
        \textbf{Ground Truth: } 3 \\
        \bottomrule
    \end{tabular}
\end{table}